\documentclass[10pt,twocolumn,letterpaper]{article}

\usepackage{wacv}              

\usepackage{graphicx}
\usepackage{amsmath}
\usepackage{amssymb}
\usepackage{booktabs}

\usepackage{makecell}    
\usepackage{colortbl}    
\usepackage{amsmath}   
\usepackage{multirow}   

\usepackage[pagebackref,breaklinks,colorlinks]{hyperref}
\usepackage{orcidlink}

\definecolor{scar}{rgb}{0.39215686, 0.58823529, 0.96078431}
\definecolor{sbicycle}{rgb}{0.39215686, 0.90196078, 0.96078431}
\definecolor{smotorcycle}{rgb}{0.11764706, 0.23529412, 0.58823529}
\definecolor{struck}{rgb}{0.31372549, 0.11764706, 0.70588235}
\definecolor{sother-vehicle}{rgb}{0.39215686, 0.31372549, 0.98039216}
\definecolor{sperson}{rgb}{1.        , 0.11764706, 0.11764706}
\definecolor{sbicyclist}{rgb}{1.        , 0.15686275, 0.78431373}
\definecolor{smotorcyclist}{rgb}{0.58823529, 0.11764706, 0.35294118}
\definecolor{sroad}{rgb}{1.        , 0.        , 1.        }
\definecolor{sparking}{rgb}{1.        , 0.58823529, 1.        }
\definecolor{ssidewalk}{rgb}{0.29411765, 0.        , 0.29411765}
\definecolor{sother-ground}{rgb}{0.68627451, 0.        , 0.29411765}
\definecolor{sbuilding}{rgb}{1.        , 0.78431373, 0.        }
\definecolor{sfence}{rgb}{1.        , 0.47058824, 0.19607843}
\definecolor{svegetation}{rgb}{0.        , 0.68627451, 0.        }
\definecolor{strunk}{rgb}{0.52941176, 0.23529412, 0.        }
\definecolor{sterrain}{rgb}{0.58823529, 0.94117647, 0.31372549}
\definecolor{spole}{rgb}{1.        , 0.94117647, 0.58823529}
\definecolor{straffic-sign}{rgb}{1.        , 0.        , 0.    }    

\definecolor{barrier}{RGB}{112,128,144}
\definecolor{bicycle}{RGB}{220,20,60}
\definecolor{bus}{RGB}{255, 127, 80}
\definecolor{car}{RGB}{255, 158, 0}
\definecolor{const. veh.}{RGB}{233, 150, 70}
\definecolor{motorcycle}{RGB}{255,61,99}
\definecolor{pedestrian}{RGB}{0,0,230}
\definecolor{traffic cone}{RGB}{47,79,79}
\definecolor{trailer}{RGB}{255,140,0}
\definecolor{truck}{RGB}{255,99,71}
\definecolor{drive. suf.}{RGB}{0,207,191}
\definecolor{other flat}{RGB}{175,0,75}
\definecolor{sidewalk}{RGB}{75,0,75}
\definecolor{terrain}{RGB}{112,180,60}
\definecolor{manmade}{RGB}{222,184,135}
\definecolor{vegetation}{RGB}{0,175,0}
\definecolor{others}{RGB}{0, 0, 0}
\usepackage[capitalize]{cleveref}
\crefname{section}{Sec.}{Secs.}
\Crefname{section}{Section}{Sections}
\Crefname{table}{Table}{Tables}
\crefname{table}{Tab.}{Tabs.}

\def\wacvPaperID{1951} 
\def\confName{WACV}
\def\confYear{2025}

\begin{document}

\title{OccLoff: Learning Optimized Feature Fusion for 3D Occupancy Prediction}

\author{Ji Zhang$^{*}$ \orcidlink{0009-0000-1759-5184} \qquad Yiran Ding\thanks{The first two authors contribute equally.} $\ $\orcidlink{0009-0005-8624-8545} \qquad Zixin Liu$\ $\orcidlink{0009-0003-2353-7847}\\
Wuhan University\\
Wuhan, Hubei Province, P.R.China. 430072\\
{\tt\small jizhang@whu.edu.cn, yrding@whu.edu.cn, liuzixin@whu.edu.cn}
 }
\maketitle

\begin{abstract}
   3D semantic occupancy prediction is crucial for finely representing the surrounding environment, which is essential for ensuring the safety in autonomous driving. Existing fusion-based occupancy methods typically involve performing a 2D-to-3D view transformation on image features, followed by computationally intensive 3D operations to fuse these with LiDAR features, leading to high computational costs and reduced accuracy. Moreover, current research on occupancy prediction predominantly focuses on designing specific network architectures, often tailored to particular models, with limited attention given to the more fundamental aspect of semantic feature learning. This gap hinders the development of more transferable methods that could enhance the performance of various occupancy models. To address these challenges, we propose OccLoff, a framework that \textbf{L}earns to \textbf{O}ptimize \textbf{F}eature \textbf{F}usion for 3D occupancy prediction. Specifically, we introduce a sparse fusion encoder with entropy masks that directly fuses 3D and 2D features, improving model accuracy while reducing computational overhead. Additionally, we propose a transferable proxy-based loss function and an adaptive hard sample weighting algorithm, which enhance the performance of several state-of-the-art methods. Extensive evaluations on the nuScenes and SemanticKITTI benchmarks demonstrate the superiority of our framework, and ablation studies confirm the effectiveness of each proposed module.
\end{abstract}

\section{Introduction}
\label{sec:intro}

Fine-grained perception of 3D scenes is crucial for ensuring the safety of autonomous vehicles. Unlike traditional 3D object detection methods~\cite{cvpr01,cvpr02,cvpr03,eccv01} and Bird’s Eye View (BEV)-based perception approaches~\cite{eccv02,arxiv01,eccv03,eccv01}, which often oversimplify the shapes of objects, 3D semantic occupancy prediction~\cite{iccv01,nips01,cvpr04,iccv02,iccv03} assigns semantic labels to every voxel in the 3D space, enabling precise perception of various irregular objects. Given that fine-grained perception better reflects the complexities of real-world driving scenarios, occupancy-based perception has emerged as a highly promising research direction.

Recent studies~\cite{iccv01,arxiv02,iral01,arxiv03} have demonstrated that leveraging the complementary characteristics of multiple sensor types, such as camera and LiDAR, for multi-modal feature fusion can substantially improve the performance of occupancy prediction. Specifically, cameras provide rich semantic information, whereas LiDARs offer precise spatial localization of objects. Despite these complementary strengths, the inherent heterogeneity between these two modalities presents a significant challenge for achieving efficient fusion. Drawing inspiration from recent advancements in camera-based perception~\cite{eccv03,cvpr05,arxiv04,cvpr07}, most existing multi-modal approaches~\cite{iccv01,iral01,arxiv03,arxiv02} perform a 2D-to-3D view transformation on image features, followed by computationally intensive 3D operations—such as 3D convolutions~\cite{iccv01,arxiv02,arxiv03} or KNN search~\cite{iral01}—to fuse these features with 3D LiDAR features. Although this approach is conceptually straightforward, it is both computationally costly and constrained by the robustness of the view transformation process~\cite{eccv02,cvpr04}.

On the other hand, existing research on occupancy-based perception primarily focuses on the detailed design of network architectures, with little attention given to the more fundamental issue of learning better feature representations for occupancy prediction. Although adjustments to network details can improve prediction performance, we argue that many techniques, which are closely tied to specific network structures, often suffer from limited transferability. Furthermore, most of the module designs in current occupancy-based approaches are either directly borrowed or minimally adapted from prior BEV perception research~\cite{nips02,icra01,arxiv05,arxiv06,eccv02,eccv03}, primarily involving the extension of 2D operations to 3D. These studies have not sufficiently explored how the inherent fine-grained nature of occupancy perception distinguishes it from BEV perception, and as a result, they have yet to fully unlocked the potential of occupancy-based perception.
\begin{figure*}[t]  
    \centering
    \includegraphics[width=\textwidth]{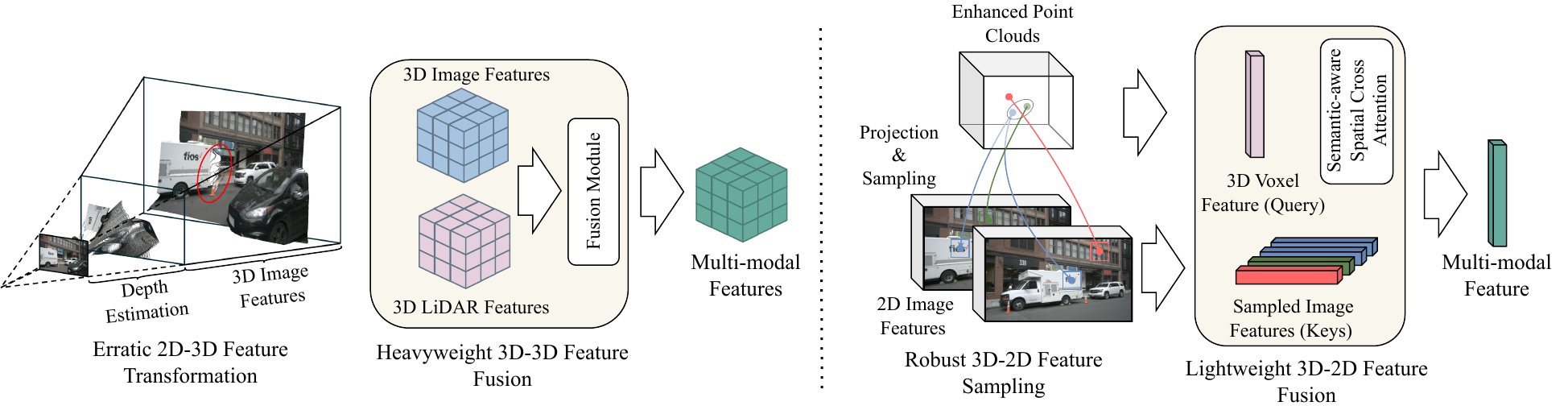}  
    \caption{\textbf{A comparison of the impact of two different image feature processing paradigms on multi-modal fusion.} The lifting paradigm requires additional modules to lift image features into 3D space before fusing them through 3D operations, which leads to high computational costs and can introduce additional noise (\eg, errors from depth estimation). In contrast, the querying from 3D to 2D approach performs feature fusion in a single step, making it more robust (see \cref{subsec:encoder} for details).}  
    \label{fig:depth_estimation_comparison}  
\end{figure*}

To address these challenges, we propose OccLoff, an occupancy prediction framework designed to facilitate more efficient multi-modal feature fusion. Rather than relying on the commonly employed lifting paradigm~\cite{eccv03}, we adopt an alternative strategy frequently used in camera-based perception—querying from 3D to 2D~\cite{cvpr06,cvpr04,eccv02,corl01}—to enable direct fusion of camera and LiDAR data. Specifically, we introduce an efficient sparse fusion encoder based on entropy masks, which leverages the prior knowledge derived from the geometric structure of point clouds to assist in feature fusion for challenging regions. Additionally, we enhance the fused data by applying sparse convolutions to multi-modal features in these difficult areas. Following the multi-modal fusion process, we incorporate multi-frame temporal information to improve robustness, and leverage CONet~\cite{iccv01} to extract fine-grained features while minimizing computational overhead. Additionally, inspired by research in fine-grained image analysis~\cite{pami01,eccv04,tip01} and metric learning~\cite{iccv04}, we introduce an occupancy proxy loss to learn more distinctive occupancy features, which significantly mitigates the class imbalance problem in occupancy semantics. Furthermore, we design an \textbf{A}daptive \textbf{H}ard \textbf{S}ample \textbf{W}eighting (AHSW) mechanism to help the model learn more effectively from complex scenes. Our approach achieves outstanding results on the nuScenes~\cite{iccv01,nips01,cvpr16} and SemanticKITTI~\cite{iccv06} benchmarks, notably improving the perception of small objects. Moreover, experiments demonstrate that our proposed learning methods are transferable and can consistently enhance the performance of other state-of-the-art occupancy prediction models.

In summary, our contributions are threefold:
(1) We propose an efficient and powerful sparse fusion encoder that better fuses LiDAR and camera features. (2) We introduce two transferable learning-based methods, the occupancy proxy loss and adaptive hard sample weighting. (3) Our framework demonstrates superior performance on the nuScenes and SemanticKITTI benchmarks.

\section{Related Work}
\subsection{3D Semantic Occupancy Prediction}
BEV-based 3D perception~\cite{eccv02,arxiv01,nips02,icra01,arxiv05,aaai01} transforms input data (\eg camera images and LiDAR point clouds) into BEV features and performs various downstream tasks in BEV space. However, these methods do not account for height information, limiting their ability to accurately reflect the actual shape of the 3D objects. In contrast, 3D semantic occupancy prediction~\cite{iccv01,nips01,cvpr04,iccv02,iral01,cvpr05} predicts semantic labels for voxels in the 3D space, representing both the geometric shape and semantic information of objects, which aligns more closely with real-world driving scenarios. SSCNet~\cite{cvpr08} was the first to introduce the task of semantic scene completion. Subsequent work has focused on occupancy-based perception using camera~\cite{cvpr04,iccv03,cvpr06,cvpr09,arxiv08}, LiDAR~\cite{iccv01,arxiv07}, or multi-modal inputs ~\cite{iccv01,iral01,arxiv03,arxiv02}. Among these, multi-modal methods combine the strengths of various sensors, offering robust performance across diverse lighting and weather conditions, and demonstrating significant potential.

\begin{figure*}[t]  
    \centering
    \includegraphics[width=\textwidth]{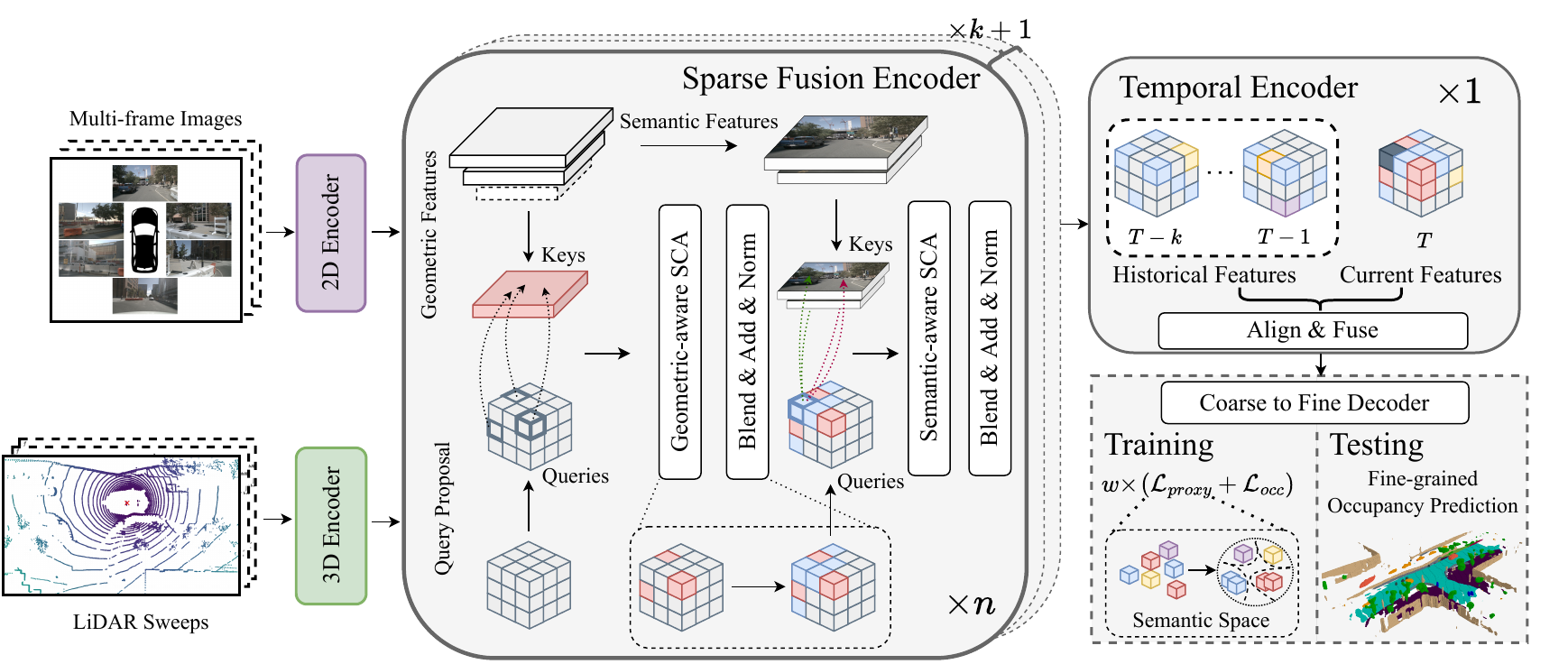}  
    \caption{\textbf{Our OccLoff framework.} The sparse fusion encoder first performs query proposal through an entropy mask, then fuses the selected LiDAR features with surrounding multi-scale image features. \textbf{SCA} represents \textbf{S}patial \textbf{C}ross \textbf{A}ttention, where the geometric-aware SCA fuses low-resolution deep features, and the semantic-aware SCA fuses high-resolution shallow features. The temporal encoder integrates multi-frame features to enhance robustness. During the training phase, each sample is weighted based on its difficulty, and the occupancy proxy loss helps obtain more distinctive occupancy features. See \cref{sec:method} for details.}  
    \label{fig:architecture}  
\end{figure*}

In camera-based surrounding perception research, two main approaches have been established. One approach~\cite{arxiv05,aaai01,iclr01,aaai02} follows the lifting paradigm introduced in LSS~\cite{eccv03}, which explicitly predicts depth to project image features into a higher-dimensional space. The other approach~\cite{cvpr04,aaai03,eccv05,cvpr10}, derived from frameworks such as BEVFormer~\cite{eccv02} and DETR3D~\cite{corl01}, employs learnable 3D queries and cross-attention mechanisms~\cite{iclr02,nips03} to extract information from image features. Since LiDAR features inherently possess a 3D spatial structure, using the lifting paradigm for dimensional alignment during fusion with camera features is a natural choice. Consequently, nearly all multi-modal occupancy-based perception studies~\cite{iccv01,arxiv02,iral01,arxiv03} adopt the lifting paradigm to process image features. However, we demonstrate that directly using LiDAR features as queries simplifies the process by eliminating the need for additional modules and reducing noise introduced by view transformation, resulting in greater simplicity and robustness, as illustrated in \cref{fig:depth_estimation_comparison}.

\subsection{Fine-grained Representation Learning}
In the domain of fine-grained image analysis, numerous studies~\cite{tip01,pami01,eccv04,cvpr11} have demonstrated that as the granularity of analysis increases, even when the core problem remains constant (e.g., learning hash code representations for images of major categories or subcategories), a greater variety of representation learning techniques is necessary to capture more distinctive features. Simply applying traditional coarse-grained methods proves insufficient for optimal results. We observe that perception methods based on occupancy or BEV face analogous challenges. Although Occ-based methods require a more detailed understanding of the scene, current approaches primarily involve making slight modifications to existing model architectures originally developed for BEV perception, without fundamentally addressing how to learn more distinctive features. To address this limitation, we propose an occupancy proxy loss and an adaptive hard sample weighting mechanism, both of which enable Occ-based models to learn more refined features, thereby enhancing perception performance.

\section{Method}
\label{sec:method}
In occupancy prediction, the multi-modal data at time \( t \) could be defined as \( I_t=\left\{C_t^1,\ldots,C_t^N,L_t\right\} \), where \( C \) represents multi-view images and \( L \) denotes LiDAR point clouds. To predict the output occupancy \( O_t \in \{s_0, s_1, \ldots, s_{N_{cls}}\}^{H \times W \times Z} \), we use \( I_t \) along with data from the preceding \( k \) frames \( I_{t-1}, \ldots, I_{t-k} \). Here, \( s_0 \) indicates the empty voxel grid, \( N_{cls} \) is the total number of semantic categories, \( H \times W \) corresponds to the dimensions of the BEV plane, and \( Z \) represents the height dimension. Our OccLoff framework can be divided into two modules: the occupancy feature encoder module (see \cref{subsec:encoder}) and distinctive feature learning module(see \cref{subsec:feature learning}).
\subsection{Occupancy Feature Encoder}
\label{subsec:encoder}
For the multi-modal data at time frame $t$, we first utilize a 2D encoder~\cite{cvpr12,cvpr13} and a 3D encoder~\cite{cvpr14,sensors01} to process the images and LiDAR point clouds, obtaining multi-view, multi-level image features \( F_{C}^{t,i,l} \in \mathbb{R}^{D \times X_l \times Y_l} \) and LiDAR features \( F_{L}^{t} \in \mathbb{R}^{D \times \frac{H}{S} \times \frac{W}{S} \times \frac{Z}{S}} \), where \( D \) represents the feature dimension, \( S \) is the upsampling ratio~\cite{iccv01}, $i$ is the camera view index, and \( l \) denotes the feature level. At a coarse granularity, we design an entropy-based query proposal mechanism (the entropy mask) to select challenging voxels for feature fusion, reducing the computational load.

\noindent \textbf{Entropy-based Query Proposal.}
Given the inherent sparsity of 3D space, occupancy sparsification enables feature optimization to concentrate solely on challenging regions, thereby reducing computational load. Previous works~\cite{cvpr06,nips01,cvpr04} generally assume that occupied voxels require additional attention, which is not always the case. The performance of occupancy prediction model could be affected by difficult regions such as the edges of small objects, whereas for many large objects, classifying occupied voxels is relatively straightforward. Inspired by this observation, before each spatial cross attention module, we employ a naive classification head to generate coarse predictions based on current 3D features, followed by computing the entropy of the resulting probability distribution:
\begin{equation}
  H^{t,v} = - \sum_{c=0}^{N_{cls}} P_{c}^{t,v} \log P_{c}^{t,v},
  \label{eq:entropy}
\end{equation}
where $v$ represents a certain voxel. We subsequently select only the top-\(K\) percent of voxels for feature fusion:
\begin{equation}
\mathcal{Q}^{t} = \{ v \mid H^{t,v} \in \text{top-}K\% \}.
    \label{eq:topk}
\end{equation}

\noindent \textbf{Geometric-aware Spatial Cross Attention.} After query proposal, we obtain the entropy mask \( \mathcal{Q}_{G-SCA}^t \), which is used to filter the current 3D features \( F_{3d}^t \) (equivalent to LiDAR features \( F_L^t \) in the first sparse fusion encoder layer), yielding the query features \( F_{\mathcal{Q}}^t \) to be used. For these queries, we apply standard spatial cross attention (SCA) ~\cite{cvpr04,eccv02,iccv02} to fuse the top-level camera features \( F_{Cam}^{t,L} \), extracting geometric information. Specifically, for a certain voxel $v$, we project the central coordinates \( p_v \) and use deformable attention~\cite{iclr02} to sample and fuse the camera features:
\begin{equation}
\mathrm{DA}(z, p, x) =
\sum_{m=1}^{N_{h}} W_{m}
\sum_{k=1}^{N_{k}} A_{mk} \cdot W'_m x(p + \Delta p_{mk}).
\label{eq:deformable}
\end{equation}
Here, \( z \) and \( x \) represent the queries and keys, respectively, \( W_m \) and \( W_m' \) are the weight matrices, \( A_{mk} \) denotes the attention weight, \( p \) is the 2D reference point, and \( \Delta p_{mk} \) represents the offset. Due to the overlapping fields of view between cameras, a single 3D point may be projected into the views of multiple cameras. We define the set of cameras capturing the 3D query points as \( V_{hit} \). The Geometric-aware SCA is expressed as:
\begin{equation}
F_{\mathrm{G-SCA}}^{t,v} = \frac{1}{|V_{\mathrm{hit}}|} \sum_{i \in V_{hit}} \mathrm{DA}(F_{\mathcal{Q}}^{t,v}, \mathcal{P}(p_v,i), F_{\mathrm{Cam}}^{t,i,L}),
    \label{eq:g-sca}
\end{equation}
where \( F_{\mathcal{Q}}^{t,v} \) represents the 3D feature of the current voxel \( v \) to be enhanced, and \( \mathcal{P}(\cdot, \cdot) \) denotes the projection function. After this, we apply sparse convolution~\cite{sensors01} to blend \( F_{\mathrm{G-SCA}}^{t} \), and then add and normalize it with \( F_{3d}^{t} \), yielding multi-modal features with enhanced geometric information.

\noindent \textbf{Semantic-aware Spatial Cross Attention.}
After integrating the geometric information contained in the deep feature maps, we use the geometric shape of the point cloud as a prior guide and employ the updated \( F_{3d}^t \) along with a \textbf{naive cross-attention mechanism}~\cite{nips03} to extract richer semantic features from the shallow feature maps. Specifically, we again use query proposal to obtain an entropy mask \( \mathcal{Q}_{S-SCA}^t \), which is applied to \( F_{3d}^t \) to obtain the query features \( F_{\mathcal{Q}}^t \). Due to the highly uneven density of the point cloud, each challenging voxel may contain a varying number of points. Let \( N_p^v \) denote the number of points within voxel \( v \). The pre-processing of point clouds is conducted as follows: for \( v \) where \( N_p^v \) exceeds \( \theta \), farthest point sampling~\cite{nips04} is applied to retain \( \theta \) points; for \( v \) with \( N_p^v \in [\tau, \theta] \), no processing is needed; and for \( v \) with \( N_p^v < \tau \), additional points are uniformly generated until the total number of points reaches \( \tau \). After pre-processing, for each voxel \( v \), all points are directly projected onto the shallow multi-layer image features, and bilinear interpolation is performed at these 2D locations to obtain the key set \( \mathbf{K} \). The query \( F_{\mathcal{Q}}^{t,v} \) is then used to fuse with the key set:
\begin{equation}
  F_{\text{S-SCA}}^{t,v} = \sum_{m=1}^{M} \Theta_m  \sum_{l\in L_{s}} \sum_{i \in V_{\text{hit}}} \sum_{j \in \mathcal{S}_{v}} A_{mlij} \cdot \Theta'_m \mathbf{K}_{v,j}^{i,l}. 
  \label{eq:S-SCA}
\end{equation}
Here, \( L_s \) represents the selected shallow feature layers, and \( \mathcal{S}_{v} \) denotes the point cloud within voxel \( v \). \( A_{mlij} \) is the attention weight, where \( \sum_{l,i,j} A_{mlij}=1 \). \(\Theta_m\) and \(\Theta'_m\) are weight matrices. Subsequently, we apply sparse convolution again, and the resulting features are added and normalized with \( F_{3d}^t \) to integrate the semantic information.

\noindent \textbf{Temporal Encoder.}
Recent breakthroughs in camera-based perception~\cite{iccv05,eccv02,cvpr04,iclr01} have demonstrated that integrating temporal information significantly enhances the performance of surrounding perception. After the n-layer sparse fusion encoder, we employ the same temporal encoder as in~\cite{cvpr04} to fuse multi-frame temporal information \( \{F_{3d}^t,F_{3d}^{t-1},...,F_{3d}^{t-k}\} \), resulting in coarse-grained multi-modal features \( F_{M}' \). In the testing phase, \( F_{M}' \) is first upsampled using the CONet decoder~\cite{iccv01} to generate fine-grained features \(F_{M}\), which are then used to directly obtain the semantic occupancy prediction results.

\begin{figure*}[t]  
    \centering
    \includegraphics[width=\textwidth]{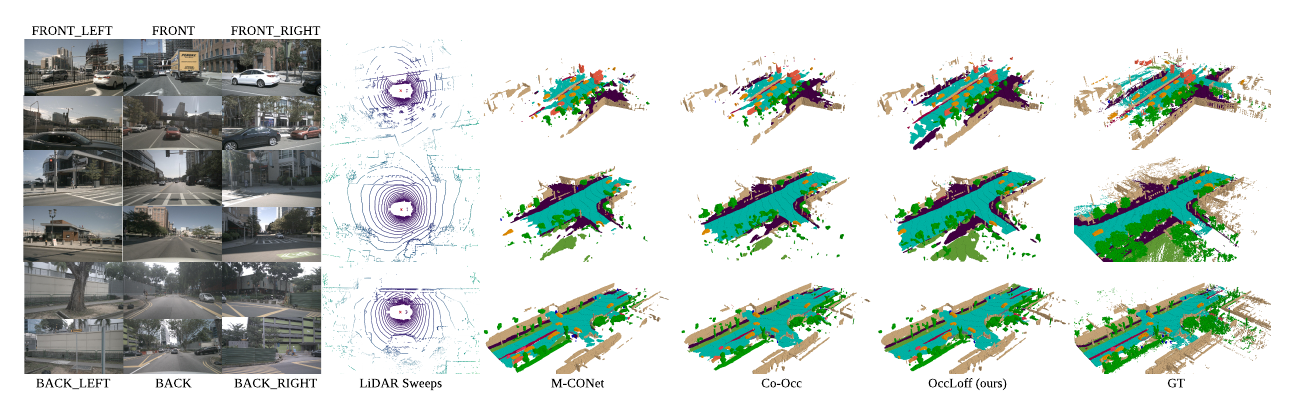}  
    \caption{\textbf{Visualization of the performance comparison between our method and existing state-of-the-art multi-modal occupancy methods on nuScenes-Occupancy~\cite{iccv01}}. Our method consistently outperforms other approaches. Better viewed when zoomed in.}  
    \label{fig:visualize}  
\end{figure*}

\subsection{Distinctive Feature Learning}
\label{subsec:feature learning}
\noindent \textbf{Occupancy Proxy Loss.}
The proxies~\cite{iccv04} are a set of learnable vectors \( \mathcal{H} \) that represent the intrinsic structure of each category in the semantic space. The number of proxies is equal to the number of semantic categories. In the semantic space, each proxy pulls the features of its corresponding category closer, while pushing away the features of all other categories, helping the model learn more distinctive features. We use the hellinger distance to measure the similarity between the feature \( F_M^v \) of a voxel \( v \) and the proxy \( h_s \) corresponding to category \( s \). Specifically, we first transform the two vectors into valid probability distributions:
\begin{equation}
 p_v = \text{Softmax}(F_M^v), \quad p_s = \text{Softmax}(h_s).
    \label{eq:softmax}
\end{equation}
Then, the following formula is used to measure the similarity:
\begin{equation}
d(F_M^v, h_s) = \frac{1}{\sqrt{2}} \left\lVert \sqrt{p_v} - \sqrt{p_s} \right\rVert_2.
    \label{eq:hellinger}
\end{equation}
Since not every sample contains all semantic categories, we divide \( \mathcal{H} \) into two parts: \( \mathcal{H}^+ \) represents proxies for categories present in sample \( V \), and \( \mathcal{H}^- \) includes proxies for categories with negative class voxels in \( V \) (note that \( \mathcal{H}^+ \) and \( \mathcal{H}^- \) are not mutually exclusive). For a category \( s \), \( V_s^+ \) represents voxels in \( V \) that belong to \( s \), and \( V_s^- = V - V_s^+ \). Our occupancy proxy loss is given by:
\begin{equation}
\begin{aligned}
\ell_{proxy}(V) &= \frac{1}{|\mathcal{H}|} 
\sum_{h_s \in \mathcal{H}} \log \left( \sum_{v \in V_{s}^+} 
e^{\alpha d(F_{M}^v,h_s)} \right)^{I(h_s \in \mathcal{H}^+)} \\
& \quad \hspace{1.6cm} \cdot \left( \sum_{v \in V_{s}^-} 
e^{-\beta d(F_{M}^v,h_s)} \right)^{I(h_s \in \mathcal{H}^-)}.
\end{aligned}
\end{equation}
Here, \( I(\cdot) \) is an indicator function, which equals 1 if the condition inside holds true, and 0 otherwise. \( \alpha \) and \( \beta \) are positive control factors. Our loss function exhibits two key advantages: first, it amplifies the gradient for difficult voxels, facilitating the extraction of more distinctive features in challenging regions; second, the gradient backpropagated from majority class voxels aids in optimizing voxels from minority classes, effectively mitigating the class imbalance issue in occupancy prediction. This is clearly illustrated by the explicit gradient of our loss function:
\begin{equation}
\frac{\partial \ell_{proxy}(V)}{\partial d(F_{M}^v, h_{s})} = 
\begin{cases}
     \frac{1}{|\mathcal{H}|} \frac{\alpha e^{\alpha d(F_{M}^v,h_{s})}}{\sum\limits_{v' \in V_{s}^+} e^{\alpha d(F_{M}^{v'},h_s)}}, & v \in V_{s}^+, \\
    \frac{1}{|\mathcal{H}|} \frac{-\beta e^{-\beta d(F_{M}^v,h_{s})}}{\sum\limits_{v' \in V_{s}^-} e^{-\beta d(F_{M}^{v'},h_s)}}, & v \in V_{s}^-.
\end{cases}
\label{eq:derivative}
\end{equation}
Specifically, the derivative of our loss function has a softmax-like form. Since the features corresponding to difficult voxels tend to be farther from the positive proxy and closer to the negative proxies, this increases the absolute value of the gradients in both cases, dynamically optimizing the prediction performance. For the proxy \( h_s \) corresponding to a minority class \( s \), the majority of voxels act as negative examples, which significantly enhances the distinctiveness of the minority class proxy (see the second part of \cref{eq:derivative}). In turn, the more distinctive \( h_s \) pulls the features of minority-class voxels closer, improving predictions for the minority class. In essence, our loss function leverages a large number of majority-class voxels to facilitate the learning of minority-class features.
In addition, we also incorporated several commonly used losses~\cite{cvpr15,cvpr05,iccv01} for occupancy prediction, which together constitute our final loss function:
\begin{equation}
\mathcal{L}_{\text{total}} = \mathcal{L}_{\text{proxy}} + \mathcal{L}_{\text{ce}} + \mathcal{L}_{\text{ls}} + \mathcal{L}_{\text{scal}}^{\text{geo}} + \mathcal{L}_{\text{scal}}^{\text{sem}}.
\end{equation}

\begin{table*}[t!]
    \centering
    
	\setlength{\tabcolsep}{0.0035\linewidth}
	\newcommand{\classfreq}[1]{{~\tiny(\semkitfreq{#1}\%)}}  %
	\centering

   \resizebox{1\linewidth}{!}{
	\begin{tabular}{l| c | c c | c c c c c c c c c c c c c c c c}
 
		\toprule
		Method
		& \makecell[c]{Input}
        & \makecell[c]{IoU}
        & \makecell[c]{mIoU}
		& \rotatebox{90}{\textcolor{barrier}{$\blacksquare$} barrier} 
		& \rotatebox{90}{\textcolor{bicycle}{$\blacksquare$} bicycle}
		& \rotatebox{90}{\textcolor{bus}{$\blacksquare$} bus} 
		& \rotatebox{90}{\textcolor{car}{$\blacksquare$} car} 
		& \rotatebox{90}{\textcolor{const. veh.}{$\blacksquare$} const. veh.} 
		& \rotatebox{90}{\textcolor{motorcycle}{$\blacksquare$} motorcycle} 
		& \rotatebox{90}{\textcolor{pedestrian}{$\blacksquare$} pedestrian} 
		& \rotatebox{90}{\textcolor{traffic cone}{$\blacksquare$} traffic cone} 
		& \rotatebox{90}{\textcolor{trailer}{$\blacksquare$} trailer} 
		& \rotatebox{90}{\textcolor{truck}{$\blacksquare$} truck} 
		& \rotatebox{90}{\textcolor{drive. suf.}{$\blacksquare$} drive. suf.} 
		& \rotatebox{90}{\textcolor{other flat}{$\blacksquare$} other flat} 
		& \rotatebox{90}{\textcolor{sidewalk}{$\blacksquare$} sidewalk} 
		& \rotatebox{90}{\textcolor{terrain}{$\blacksquare$} terrain} 
		& \rotatebox{90}{\textcolor{manmade}{$\blacksquare$} manmade} 
		& \rotatebox{90}{\textcolor{vegetation}{$\blacksquare$} vegetation} \\
		\midrule
  
  		TPVFormer~\cite{cvpr09} & C &15.3&  7.8 & 9.3  & 4.1  &  11.3 &  10.1 & 5.2  & 4.3  & 5.9 & 5.3&  6.8& 6.5 & 13.6 & 9.0  & 8.3 & 8.0  & 9.2 & 8.2 \\

            LMSCNet~\cite{3DV01} & L &27.3& 11.5 & 12.4&  4.2 & 12.8  & 12.1  & 6.2  &  4.7 & 6.2 & 6.3&  8.8&  7.2& 24.2 & 12.3  & 16.6 & 14.1  & 13.9 & 22.2 \\

            L-CONet~\cite{iccv01} & L  &30.9& 15.8 &  17.5  & 5.2  & 13.3  & 18.1  & 7.8  & 5.4  & 9.6 & 5.6& 13.2 & 13.6 & 34.9 & 21.5  & 22.4 & 21.7  & 19.2 &23.5  \\

            M-CONet~\cite{iccv01} & C\&L &29.5& 20.1 &  23.3  & 13.3  & 21.2  & 24.3  & 15.3  & 15.9  & 18.0 & 13.3 & 15.3 & 20.7 & 33.2 & 21.0 & 22.5 & 21.5 & 19.6 & 23.2  \\
            
            Co-Occ~\cite{iral01} & C\&L &30.6& 21.9 & 26.5 & 16.8 & 22.3 & \textbf{27.0} & 10.1 & 20.9 & 20.7 & 14.5 & 16.4 & 21.6 & 36.9 & 23.5 & \textbf{25.5} & 23.7 & 20.5 & 23.5 \\
            OccGen~\cite{arxiv03} &C\&L &30.3&22.0&24.9&16.4&22.5&26.1&14.0&20.1&21.6&14.6&\textbf{17.4}&21.9&35.8&\textbf{24.5}&24.7&\textbf{24.0}&20.5&23.5\\
            \textbf{OccLoff} (ours) & C\&L &\textbf{31.4}&\textbf{22.9} & 
\textbf{26.7} & \textbf{17.2} & \textbf{22.6} & 26.9 & \textbf{16.4} & \textbf{22.6} & \textbf{24.7} & \textbf{16.4} & 16.3 & \textbf{22.0} &
\textbf{37.5} & 22.3 & 25.3 &
23.9 & \textbf{21.4} & \textbf{24.2} \tabularnewline
            
		\bottomrule
	\end{tabular}}

    \caption{\textbf{Performance on the nuScenes-Occupancy validation set~\cite{iccv01}.} C and L represent camera and LiDAR, respectively.}
    \label{tab:nuscenes_occ}
\end{table*}

\noindent \textbf{Adaptive Hard Sample Weighting.}
Surrounding semantic perception models demonstrate varying performance across different scenarios~\cite{arxiv02,icra01}, with certain conditions, such as nighttime or rain, posing greater challenges. To help the model focus on these difficult samples, we dynamically weight the gradients of samples. Specifically, samples with larger losses during each epoch are given increased attention, while their importance decreases as their loss diminishes. During the initial \( \mathcal{M} \) epochs, no weighting is applied to allow the model to converge. For each sample, we track its loss across epochs as \( \mathbf{L}_{n-1} = \{\ell^{(1)}_V, \ldots, \ell^{(n-1)}_V\} \). The cumulative loss for sample \( V \) in epoch \( n \) is defined as:
\begin{equation}
\mathcal{C}_V^{(n)} = \sum_{e=1}^{n-1} \gamma^{n-1-e} \ell_V^{(e)},
\label{eq:cumulative loss}
\end{equation}
where \( 0 < \gamma < 1 \) is the scaling factor. When \( n > \mathcal{M} \), we sort all samples in descending order based on their cumulative loss and only use the top \( \mathcal{K} \) percent of samples for backpropagation. Additionally, for a specific sample \( V \), we scale its gradient using \( w_V^{(n)} \) as follows:

\begin{equation}
w_V^{(n)} = 1 + (\lambda - 1) \cdot \frac{\mathcal{C}_V^{(n)} - \mathcal{C}_{\text{min}}^{(n)}}{\mathcal{C}_{\text{max}}^{(n)} - \mathcal{C}_{\text{min}}^{(n)}}
, \quad \tilde{\ell}_V^{(n)} = w_V^{(n)} \cdot \ell_V^{(n)}.
\label{eq:weight}
\end{equation}
Here, \( \mathcal{C}_{\text{max}}^{(n)} \) and \( \mathcal{C}_{\text{min}}^{(n)} \) represent the maximum and minimum cumulative losses of the samples in the current epoch, respectively, and \( \lambda > 1 \) is the gradient amplification factor. The loss of the samples involved in training is multiplied by \( w_V^{(n)} \), while the losses of non-participating samples are computed but do not contribute to backpropagation.

\section{Experiments}
\subsection{Experimental Setup}
\noindent \textbf{Dataset and Metrics.}
Following standard practice~\cite{arxiv03,arxiv02,iral01}, we trained our proposed OccLoff model on the training sets of nuScenes-Occupancy~\cite{iccv01}, SemanticKITTI~\cite{iccv06}, and nuScenes-Occ3D~\cite{nips01}, and evaluated it on the corresponding validation sets. The annotated nuScenes datasets~\cite{iccv01,nips01,cvpr16} consist of 700 training scenes and 150 validation scenes, while SemanticKITTI~\cite{iccv06} includes 22 sequences of monocular images, LiDAR point clouds, and ground truth for semantic scene completion. We evaluated our model using two common metrics: Intersection over Union (IoU) and mean Intersection over Union (mIoU), as described in~\cite{iccv01}.

\noindent \textbf{Implementation Details.}
We adopt an experimental setup similar to that in~\cite{iccv01,iral01} to ensure a fair comparison with these methods. We train three models of different scales to evaluate the effectiveness of each proposed approach (see \cref{tab:ablate size}). Specifically, for the standard OccLoff-Base and OccLoff-Dense models (without sparse operations), we use an ImageNet~\cite{cvpr17} pretrained ResNet101~\cite{cvpr13} with FPN~\cite{cvpr12} as the 2D backbone, alongside three sparse fusion encoders. For the lightweight OccLoff-Small model, we employ an ImageNet~\cite{cvpr17} pretrained ResNet50~\cite{cvpr13} as the backbone and, following~\cite{iclr02}, apply $1\times1$ convolution to unify the channel dimensions of the feature maps from the last three stages, using only two sparse fusion encoders. The feature dimension \( D \) is set to 256, and \( K \) is set to 35 in the query proposal. The number of sampling points for deformable attention~\cite{iclr02} is set to 8, with S-SCA utilizing the output features from the penultimate and antepenultimate stages. The input image size is $1600\times900$, and 10 LiDAR sweeps are voxelized as input for the 3D encoder~\cite{cvpr14,sensors01}, with the decoder's upsampling rate \( S \) set to 2. In the temporal encoder, features from the 4 most recent frames (including the current frame) are fused. During training, we use the AdamW optimizer~\cite{iclr03} with a weight decay of 0.01 and an initial learning rate of 2e-4. The models are trained for 20 epochs on 8 A100 GPUs, with a batch size of 8.

\subsection{Main Results}
Our model achieved state-of-the-art performance across three challenging benchmarks, surpassing M-CONet~\cite{iccv01} and other multi-modal methods~\cite{arxiv02,arxiv03,iral01} in terms of IoU for most categories (it is worth noting that OccGen~\cite{arxiv03} and OccFusion~\cite{arxiv02} are concurrent works with ours), see \cref{tab:nuscenes_occ,tab:occ3d,tab:semantickitti}. Notably, our approach demonstrated significant improvements in small object categories. Specifically, on the nuScenes-Occupancy dataset~\cite{iccv01}, we achieved substantial IoU gains over M-CONet~\cite{iccv01} for the motorcycle, pedestrian, and traffic cone categories, with relative increases of 42.1\%, 37.2\%, and 23.3\%, respectively. Furthermore, our model consistently outperformed other methods in small object categories across all three benchmarks, highlighting its ability to more distinctively learn occupancy features and improve fine-grained perception of small objects.

\begin{table*}
    \centering
    
        \setlength{\tabcolsep}{0.0035\linewidth}
	\newcommand{\classfreq}[1]{{~\tiny(\semkitfreq{#1}\%)}}  %
	\centering

 \resizebox{1\linewidth}{!}{
  \begin{tabular}{c|c|c|ccccccccccccccccc}
    \toprule
    Method & Input & mIoU & 
    \rotatebox{90}{\textcolor{others}{$\bullet$} others} &
    \rotatebox{90}{\textcolor{barrier}{$\bullet$} barrier} & \rotatebox{90}{\textcolor{bicycle}{$\bullet$} bicycle} & \rotatebox{90}{\textcolor{bus}{$\bullet$} bus} & \rotatebox{90}{\textcolor{car}{$\bullet$} car} & \rotatebox{90}{\textcolor{const. veh.}{$\bullet$} const. veh.} & \rotatebox{90}{\textcolor{motorcycle}{$\bullet$} motorcycle} & \rotatebox{90}{\textcolor{pedestrian}{$\bullet$} pedestrian} & \rotatebox{90}{\textcolor{traffic cone}{$\bullet$} traffic cone} & \rotatebox{90}{\textcolor{trailer}{$\bullet$} trailer} & \rotatebox{90}{\textcolor{truck}{$\bullet$} truck} & \rotatebox{90}{\textcolor{drive. suf.}{$\bullet$} drive. surf.} & \rotatebox{90}{\textcolor{other flat}{$\bullet$} other flat} & \rotatebox{90}{\textcolor{sidewalk}{$\bullet$} sidewalk} & \rotatebox{90}{\textcolor{terrain}{$\bullet$} terrain} & \rotatebox{90}{\textcolor{manmade}{$\bullet$} manmade} & \rotatebox{90}{\textcolor{vegetation}{$\bullet$} vegetation} \\
     \midrule 

    OccFormer ~\cite{iccv03}  & C & 21.93 & 5.94 & 30.29 & 12.32 & 34.40 & 39.17 & 14.44 & 16.45 & 17.22 & 9.27 & 13.90 & 26.36 & 50.99 & 30.96 & 34.66 & 22.73 & 6.76 & 6.97 \\
    CTF-Occ ~\cite{nips01} & C & 28.53 & 8.09 & 39.33 & 20.56 & 38.29 & 42.24 & 16.93 & 24.52 & 22.72 & 21.05 & 22.98 & 31.11 & 53.33 & 33.84 & 37.98 & 33.23 & 20.79 & 18.00 \\
    RenderOcc ~\cite{icra02}  & C & 26.11 & 4.84 & 31.72 & 10.72 & 27.67 & 26.45 & 13.87 & 18.20 & 17.67 & 17.84 & 21.19 & 23.25 & 63.20 & 36.42 & 46.21 & 44.26 & 19.58 & 20.72 \\
    PanoOcc ~\cite{cvpr04} & C & 42.13 & 11.67 & 50.48 & 29.64 & 49.44 & 55.52 & 23.29 & 33.26 & 30.55 & 30.99 & 34.43 & 42.57 & 83.31 & 44.23 & 54.40 & 56.04 & 45.94 & 40.40 \\

    OccFusion ~\cite{arxiv02} & C\&L\&R & 46.67 & 12.37 & 50.33 & 31.53 & \textbf{57.62} & 58.81 & 33.97 & 41.00 & 47.18 & 29.67 & \textbf{42.03} & 48.04 & 78.39 & 35.68 & 47.26 & 52.74 & 63.46 & \textbf{63.30} \\
    
    \textbf{OccLoff} (ours) & C\&L & \textbf{49.36} & \textbf{13.26} & \textbf{53.72} & \textbf{33.20} & 55.21 & \textbf{58.94} & \textbf{34.26} & \textbf{43.13} & \textbf{49.28} &\textbf{35.61} & 41.44 & \textbf{48.78}& \textbf{83.72} & \textbf{44.68} & \textbf{57.33} & \textbf{60.15} & \textbf{63.89} & 62.45 \\
    
    \bottomrule
  \end{tabular}}

    \caption{\textbf{3D semantic occupancy prediction results on nuScenes-Occ3D~\cite{nips01} validation set.} \emph{C}, \emph{L}, \emph{R} represent camera, LiDAR, and radar, respectively.}
    \label{tab:occ3d}
\end{table*}

\begin{table*}[t!]
    \centering
    
\footnotesize
\setlength{\tabcolsep}{0.004\linewidth}
\centering
\begin{tabular}{l|c|c| c c c c c c c c c c c c c c c c c c c}
    \toprule
    Method
        & \makecell[c]{Modality}
     & mIoU
    & \rotatebox{90}{\textcolor{sroad}{$\blacksquare$} road}
    & \rotatebox{90}{\textcolor{ssidewalk}{$\blacksquare$} sidewalk}
    & \rotatebox{90}{\textcolor{sparking}{$\blacksquare$} parking} 
    & \rotatebox{90}{\textcolor{sother-ground}{$\blacksquare$} other-grnd} 
    & \rotatebox{90}{\textcolor{sbuilding}{$\blacksquare$}  building} 
    & \rotatebox{90}{\textcolor{scar}{$\blacksquare$}  car} 
    & \rotatebox{90}{\textcolor{struck}{$\blacksquare$}  truck} 
    & \rotatebox{90}{\textcolor{sbicycle}{$\blacksquare$}  bicycle} 
    & \rotatebox{90}{\textcolor{smotorcycle}{$\blacksquare$} motorcycle} 
    & \rotatebox{90}{\textcolor{sother-vehicle}{$\blacksquare$}  other-veh.} 
    & \rotatebox{90}{\textcolor{svegetation}{$\blacksquare$} vegetation} 
    & \rotatebox{90}{\textcolor{strunk}{$\blacksquare$}  trunk} 
    & \rotatebox{90}{\textcolor{sterrain}{$\blacksquare$} terrain} 
    & \rotatebox{90}{\textcolor{sperson}{$\blacksquare$}  person} 
    & \rotatebox{90}{\textcolor{sbicyclist}{$\blacksquare$}  bicyclist} 
    & \rotatebox{90}{\textcolor{smotorcyclist}{$\blacksquare$}  motorcyclist.} 
    & \rotatebox{90}{\textcolor{sfence}{$\blacksquare$} fence} 
    & \rotatebox{90}{\textcolor{spole}{$\blacksquare$} pole} 
    & \rotatebox{90}{\textcolor{straffic-sign}{$\blacksquare$} traf.-sign} 
    \\
    \midrule

VoxFormer~\cite{cvpr06}&C  & 12.35 & 54.76 & 26.35 & 
    15.50 & \textbf{0.70} &17.65 & 25.79 & 5.63 & 0.59 & 0.51 & 3.77 & 24.39 & 5.08 & 29.96 & 1.78 & 3.32 & 0.00 & 7.64& 7.11 & 4.18 \\ 
    JS3C-Net~\cite{aaai04}&C  &10.31& 50.49& 23.74 & 11.94 & 0.07& 15.03  &24.65 & 4.41 & 0.00 & 0.00& 6.15  &18.11&4.33 & 26.86 &0.67  &0.27 & 0.00 & 3.94 & 3.77 &1.45   \\
MonoScene~\cite{cvpr05}&C  & 11.08 & 56.52 & 26.72 &  14.27& 0.46 & 14.09 & 23.26 & 6.98 & 0.61 & 0.45 & 1.48& 17.89 & 2.81 & 29.64      & 1.86 & 1.20 & 0.00 & 5.84 & 4.14 & 2.25  \\
    LMSCNet~\cite{3DV01} &L &6.70&40.68& 18.22 &4.38& 0.00& 10.31& 18.33 &0.00& 0.0& 0.0& 0.0& 13.66& 0.02 &20.54& 0.00& 0.00& 0.00& 1.21& 0.00& 0.00\\

    OccFusion~\cite{arxiv02}&C\&L& 21.64&65.73& 36.46& \textbf{21.35} &0.00 &39.32 &45.72 &19.41 &2.00 &3.39& 7.92& \textbf{41.21} &19.18 &\textbf{46.28} &2.33& 3.07& 0.00& 15.74& \textbf{27.60}& 14.36\\
    M-CONet~\cite{iccv01}&C\&L  & 17.25 & 58.00 & 29.44 &19.33  & 0.02 & 27.85 &33.25 & 19.46 &0.66 &2.2  & 5.99 & 32.56 &16.01 &37.31 &2.07 & 2.38 & 0.00 & 12.25 &14.88 & 14.09  \\
    
    \textbf{OccLoff} (Ours)&C\&L & \textbf{22.62}  &\textbf{66.25}  &\textbf{43.51} & 21.07 & 0.57 &\textbf{41.23}  & \textbf{46.44} & \textbf{20.38}&\textbf{2.08}&\textbf{3.91}&\textbf{8.72}&41.20&\textbf{20.06}&46.21& \textbf{3.88} & \textbf{4.35}& 0.00 & \textbf{15.86}& \textbf{27.60}& \textbf{16.47}   \\
    \bottomrule
\end{tabular}
    \caption{\textbf{3D Semantic occupancy prediction results on SemanticKITTI~\cite{iccv06} validation set.} The \emph{C} and \emph{L} denote camera and LiDAR.}
    \label{tab:semantickitti}  
\end{table*}

\subsection{Ablation Studies}
We conduct extensive ablation experiments on the \textbf{nuScenes-Occupancy benchmark}~\cite{iccv01} to demonstrate the effectiveness of each proposed module and the transferability of our learning methods.

\noindent \textbf{Ablation on Main Components.}
The contribution of each module is summarized in \cref{tab:ablate components}. While removing any module degrades the overall performance, our model consistently maintains state-of-the-art results. As in previous studies~\cite{eccv02,cvpr04}, incorporating temporal information provides a minor performance improvement. The semantic-aware spatial cross attention (S-SCA), a core module, extracts fine-grained semantic information from multi-layer image features. Excluding this module results in a significant accuracy decline, with a 2.6 mIoU decrease. The geometric-aware spatial cross attention (G-SCA) captures coarse-grained geometric features, which guide the fusion of semantic information from images, and is also crucial to model performance. Both of our proposed learning methods aid in model convergence. Although the mIoU improvement from adaptive hard sample weighting (AHSW) is relatively smaller, it consistently accelerates convergence (see \cref{tab:transfer}).

\begin{table}[ht]
    \centering
    
    \begin{tabular}{c c|c c c |c}
        \toprule
        P.L. & AHSW & G-SCA&S-SCA&Temp.E.& mIoU \\
        \midrule
        \checkmark &\checkmark &\checkmark &\checkmark & & 22.4 \\
        \checkmark & \checkmark &\checkmark & &\checkmark & 20.3 \\
        \checkmark & \checkmark &  &\checkmark &\checkmark & 22.0 \\
        \checkmark &  & \checkmark & \checkmark&\checkmark &22.1 \\
  & \checkmark & \checkmark & \checkmark&\checkmark &21.5 \\
  &  & \checkmark & \checkmark&\checkmark &21.2 \\
  \checkmark & \checkmark & \checkmark & \checkmark&\checkmark &\textbf{22.9} \\
        \bottomrule
    \end{tabular}
    \caption{\textbf{Ablation study on proposed components.} P.L. represents the occupancy proxy loss, AHSW stands for adaptive hard sample weighting, G-SCA and S-SCA refer to geometric-aware and semantic-aware spatial cross attention, respectively, and Temp.E. denotes the temporal encoder~\cite{cvpr04}.}
    \label{tab:ablate components}
\end{table}
\noindent \textbf{Ablation Studies on Hyper-parameters.}
Several modules in our architecture, including the sparse fusion encoder, occupancy proxy loss, and adaptive hard sample weighting, involve hyperparameters. However, experiments indicate that our model is generally robust to hyperparameter selection (see \cref{tab:ablate hyper parameters}). In AHSW, slightly increasing the number of warm-up epochs \( \mathcal{M} \) and the sample proportion per epoch \( \mathcal{K} \) enhances model performance, while an excessively large weighting factor \( \lambda \) disrupts convergence and slightly reduces effectiveness. A smaller \( \gamma \) allows the model to focus on difficult samples from recent epochs, facilitating learning from more challenging examples. In the proxy loss, increasing \( \beta \) proves more beneficial for convergence than increasing \( \alpha \), likely because a larger \( \beta \) amplifies the gradient contribution from minority class proxies, improving feature distinctiveness. For point cloud pre-processing in S-SCA, increasing the number of points offers marginal performance gains but introduces additional computational costs, making it not worth the trade-off.
\begin{table}[ht]
    \centering
    
    \begin{tabular}{c c|c c| c c c c |c}
        \toprule
        $\tau$&$\theta$&$\alpha$ & $\beta$ & $\mathcal{M}$ & $\mathcal{K}$ & $\lambda$ & $\gamma$ & mIoU \\
        \midrule
        \multicolumn{9}{c}{\textit{Ablation on AHSW}} \\
        5&20&/ & / & 5 & 50 & 5 & 0.5  & 21.1 \\
        5&20&/ & / & 10 & 70 & 5 & 0.5 & \textbf{21.5} \\
        5&20&/ & / & 10 & 70 & 10 & 0.5 & 21.2 \\
        5&20&/ & / & 10 & 70 & 5 & 0.8 & 20.9 \\
        \midrule
        \multicolumn{9}{c}{\textit{Ablation on P.L.}} \\
        5&20&6 & 6  & 10 & 70 & 5 & 0.5 & 22.4 \\
        5&20&12 & 6 & 10 & 70 & 5 & 0.5 & 22.4 \\
        5&20&6 & 12 & 10 & 70 & 5 & 0.5 & \textbf{22.9} \\
        \midrule
        \multicolumn{9}{c}{\textit{Ablation on P.S.}} \\
        5&20&6 & 12 & 10 & 70 & 5 & 0.5 & 22.9 \\
        10&30&6 & 12 & 10 & 70 & 5 & 0.5 & \textbf{23.1} \\
        \bottomrule
    \end{tabular}
    \caption{\textbf{Ablation study on hyper-parameters.} P.S. represents the point cloud sampling technique used in S-SCA. Our model is generally robust to hyperparameter changes.}
    \label{tab:ablate hyper parameters}
\end{table}

\noindent \textbf{Ablation on Model Efficiency.}
We trained three versions of OccLoff: Small (S), Base (B), and Dense (D). The Base version serves as our standard model for performance comparison, the Small version is a lightweight variant, and the Dense version omits the query proposal mechanism to highlight the effectiveness of sparse operations (see \cref{tab:ablate size}). Incorporating FPN~\cite{cvpr12} and multi-layer stacked sparse fusion encoders slightly enhances model performance. Notably, the sparse versions not only improve computational efficiency but also boost mIoU, likely due to the query proposal mechanism, which repeatedly selects new challenging regions between the two SCA layers, allowing the model to dynamically learn more effective features. Our OccLoff-S model achieves state-of-the-art accuracy with extremely low latency, demonstrating the effectiveness of our architectural design.
\begin{table}[ht]
    \centering
    \begin{tabular}{l|c|c|c|c}
        \toprule
        Method &Backbone & Memory & Latency &  mIoU \\
        \midrule
        OccLoff-S &R50& \textbf{ 6.1 G} & \textbf{198 ms}  & 22.2 \\
        \textbf{OccLoff-B} &R101+FPN&  7.9 G & 285 ms  & \textbf{22.9} \\
        OccLoff-D &R101+FPN& 16.4 G & 487 ms  & 22.6 \\
        \bottomrule
    \end{tabular}
    \caption{\textbf{Ablation study on the model backbone and sparse encoder.} Memory represents the GPU memory usage during the testing phase, with tests conducted on a single A100 GPU. Note that sparse operations can balance both efficiency and accuracy.}
    \label{tab:ablate size}
\end{table}

\noindent \textbf{The Transferability of Learning Methods.}
Existing occupancy-based perception research often focuses heavily on model architecture design while neglecting the fine-grained nature of the occupancy task itself. In contrast, our occupancy proxy loss and adaptive hard sample weighting (AHSW) are designed to help the model learn more distinctive features from two key perspectives. First, occupancy categories exhibit a long-tail distribution, where minority-class voxels contribute less to the gradient, leading to poorer prediction performance. Second, difficult samples may be critical for model convergence and should receive special attention. Our occupancy proxy loss enables the gradient from majority-class voxels to propagate towards minority-class proxies, refining the features of minority-class voxels, see \cref{eq:derivative}. Additionally, AHSW applies adaptive weighting to challenging samples without disrupting the training process, facilitating model convergence. As shown in \cref{tab:transfer}, our proposed learning methods improve the performance of several state-of-the-art models, demonstrating their effectiveness. While the accuracy improvement from AHSW is modest, it consistently accelerates convergence. Given the computational intensity of training occupancy prediction models, achieving better models in less time offers significant potential.

\begin{table}[ht]
    \centering
    \begin{tabular}{c|c c|c|c}
        \toprule
        Model & P.L. & AHSW & mIoU&Epochs \\
        \midrule
        \multirow{2}{*}{M-CONet~\cite{iccv01}} 
        & \checkmark &  & 20.9 (+0.8)&24 \\
        & \checkmark &\checkmark & \textbf{21.1}(+1.0)&\textbf{16} \\
        \midrule
        \multirow{2}{*}{Co-Occ~\cite{iral01}} 
        & \checkmark &   & \textbf{22.3} (+0.4)&24 \\
        & \checkmark & \checkmark & 22.2 (+0.3) & \textbf{18} \\
        \bottomrule
    \end{tabular}
    \caption{\textbf{Ablation study on the transferability of the learning methods. }Note that AHSW can accelerate model convergence. }
    \label{tab:transfer}
\end{table}
\section{Conclusion}
In this paper, we propose OccLoff, a framework that learns to optimize feature fusion for 3D occupancy prediction, aiming to effectively enhance occupancy-based perception from the perspective of fine-grained feature learning. Our efficient sparse fusion encoder enables the effective fusion of LiDAR and camera features, and we introduce two transferable learning-based methods to improve the performance of other occupancy prediction models. Extensive experiments on nuScenes-Occupancy, nuScenes-Occ3D, and SemanticKITTI demonstrate the effectiveness of our framework. Additionally, our learning-based methods further improve two state-of-the-art baselines, highlighting the potential of our approach.

{\small
\bibliographystyle{ieee_fullname}
\bibliography{OccLoff}

\begin{thebibliography}{10}\itemsep=-1pt

\bibitem{iccv06}
Jens Behley, Martin Garbade, Andres Milioto, Jan Quenzel, Sven Behnke, Cyrill Stachniss, and J{\"{u}}rgen Gall.
\newblock Semantickitti: {A} dataset for semantic scene understanding of lidar sequences.
\newblock In {\em 2019 {IEEE/CVF} International Conference on Computer Vision, {ICCV} 2019, Seoul, Korea (South), October 27 - November 2, 2019}, pages 9296--9306, 2019.

\bibitem{cvpr15}
Maxim Berman, Amal~Rannen Triki, and Matthew~B. Blaschko.
\newblock The lov{\'{a}}sz-softmax loss: {A} tractable surrogate for the optimization of the intersection-over-union measure in neural networks.
\newblock In {\em 2018 {IEEE} Conference on Computer Vision and Pattern Recognition, {CVPR} 2018, Salt Lake City, UT, USA, June 18-22, 2018}, pages 4413--4421, 2018.

\bibitem{cvpr16}
Holger Caesar, Varun Bankiti, Alex~H. Lang, Sourabh Vora, Venice~Erin Liong, Qiang Xu, Anush Krishnan, Yu Pan, Giancarlo Baldan, and Oscar Beijbom.
\newblock nuscenes: {A} multimodal dataset for autonomous driving.
\newblock In {\em 2020 {IEEE/CVF} Conference on Computer Vision and Pattern Recognition, {CVPR} 2020, Seattle, WA, USA, June 13-19, 2020}, pages 11618--11628, 2020.

\bibitem{cvpr05}
Anh{-}Quan Cao and Raoul de Charette.
\newblock Monoscene: Monocular 3d semantic scene completion.
\newblock In {\em {IEEE/CVF} Conference on Computer Vision and Pattern Recognition, {CVPR} 2022, New Orleans, LA, USA, June 18-24, 2022}, pages 3981--3991, 2022.

\bibitem{cvpr01}
Ming{-}Fang Chang, John Lambert, Patsorn Sangkloy, Jagjeet Singh, Slawomir Bak, Andrew Hartnett, De Wang, Peter Carr, Simon Lucey, Deva Ramanan, and James Hays.
\newblock Argoverse: 3d tracking and forecasting with rich maps.
\newblock In {\em {IEEE} Conference on Computer Vision and Pattern Recognition, {CVPR} 2019, Long Beach, CA, USA, June 16-20, 2019}, pages 8748--8757, 2019.

\bibitem{tip01}
Zhen{-}Duo Chen, Xin Luo, Yongxin Wang, Shanqing Guo, and Xin{-}Shun Xu.
\newblock Fine-grained hashing with double filtering.
\newblock {\em {IEEE} Trans. Image Process.}, 31:1671--1683, 2022.

\bibitem{cvpr11}
Zhen-Duo Chen, Li-Jun Zhao, Zi-Chao Zhang, Xin Luo, and Xin-Shun Xu.
\newblock Characteristics matching based hash codes generation for efficient fine-grained image retrieval.
\newblock In {\em Proceedings of the IEEE/CVF Conference on Computer Vision and Pattern Recognition (CVPR)}, pages 17273--17281, 2024.

\bibitem{eccv04}
Quan Cui, Qing{-}Yuan Jiang, Xiu{-}Shen Wei, Wu{-}Jun Li, and Osamu Yoshie.
\newblock Exchnet: {A} unified hashing network for large-scale fine-grained image retrieval.
\newblock In {\em Computer Vision - {ECCV} 2020 - 16th European Conference, Glasgow, UK, August 23-28, 2020, Proceedings, Part {III}}, volume 12348, pages 189--205, 2020.

\bibitem{cvpr17}
Jia Deng, Wei Dong, Richard Socher, Li{-}Jia Li, Kai Li, and Li Fei{-}Fei.
\newblock Imagenet: {A} large-scale hierarchical image database.
\newblock In {\em 2009 {IEEE} Computer Society Conference on Computer Vision and Pattern Recognition {(CVPR} 2009), 20-25 June 2009, Miami, Florida, {USA}}, pages 248--255, 2009.

\bibitem{cvpr02}
Andreas Geiger, Philip Lenz, and Raquel Urtasun.
\newblock Are we ready for autonomous driving? the {KITTI} vision benchmark suite.
\newblock In {\em 2012 {IEEE} Conference on Computer Vision and Pattern Recognition, Providence, RI, USA, June 16-21, 2012}, pages 3354--3361, 2012.

\bibitem{cvpr13}
Kaiming He, Xiangyu Zhang, Shaoqing Ren, and Jian Sun.
\newblock Deep residual learning for image recognition.
\newblock In {\em 2016 {IEEE} Conference on Computer Vision and Pattern Recognition, {CVPR} 2016, Las Vegas, NV, USA, June 27-30, 2016}, pages 770--778, 2016.

\bibitem{arxiv06}
Junjie Huang and Guan Huang.
\newblock Bevdet4d: Exploit temporal cues in multi-camera 3d object detection.
\newblock {\em CoRR}, abs/2203.17054, 2022.

\bibitem{arxiv05}
Junjie Huang, Guan Huang, Zheng Zhu, and Dalong Du.
\newblock Bevdet: High-performance multi-camera 3d object detection in bird-eye-view.
\newblock {\em CoRR}, abs/2112.11790, 2021.

\bibitem{cvpr09}
Yuanhui Huang, Wenzhao Zheng, Yunpeng Zhang, Jie Zhou, and Jiwen Lu.
\newblock Tri-perspective view for vision-based 3d semantic occupancy prediction.
\newblock In {\em {IEEE/CVF} Conference on Computer Vision and Pattern Recognition, {CVPR} 2023, Vancouver, BC, Canada, June 17-24, 2023}, pages 9223--9232, 2023.

\bibitem{aaai03}
Yanqin Jiang, Li Zhang, Zhenwei Miao, Xiatian Zhu, Jin Gao, Weiming Hu, and Yu{-}Gang Jiang.
\newblock Polarformer: Multi-camera 3d object detection with polar transformer.
\newblock In {\em Thirty-Seventh {AAAI} Conference on Artificial Intelligence, {AAAI} 2023, Thirty-Fifth Conference on Innovative Applications of Artificial Intelligence, {IAAI} 2023, Thirteenth Symposium on Educational Advances in Artificial Intelligence, {EAAI} 2023, Washington, DC, USA, February 7-14, 2023}, pages 1042--1050, 2023.

\bibitem{iclr03}
Diederik~P. Kingma and Jimmy Ba.
\newblock Adam: {A} method for stochastic optimization.
\newblock In {\em 3rd International Conference on Learning Representations, {ICLR} 2015, San Diego, CA, USA, May 7-9, 2015, Conference Track Proceedings}, 2015.

\bibitem{aaai02}
Yinhao Li, Han Bao, Zheng Ge, Jinrong Yang, Jianjian Sun, and Zeming Li.
\newblock Bevstereo: Enhancing depth estimation in multi-view 3d object detection with temporal stereo.
\newblock In {\em Thirty-Seventh {AAAI} Conference on Artificial Intelligence, {AAAI} 2023, Thirty-Fifth Conference on Innovative Applications of Artificial Intelligence, {IAAI} 2023, Thirteenth Symposium on Educational Advances in Artificial Intelligence, {EAAI} 2023, Washington, DC, USA, February 7-14, 2023}, pages 1486--1494, 2023.

\bibitem{aaai01}
Yinhao Li, Zheng Ge, Guanyi Yu, Jinrong Yang, Zengran Wang, Yukang Shi, Jianjian Sun, and Zeming Li.
\newblock Bevdepth: Acquisition of reliable depth for multi-view 3d object detection.
\newblock In {\em Thirty-Seventh {AAAI} Conference on Artificial Intelligence, {AAAI} 2023, Thirty-Fifth Conference on Innovative Applications of Artificial Intelligence, {IAAI} 2023, Thirteenth Symposium on Educational Advances in Artificial Intelligence, {EAAI} 2023, Washington, DC, USA, February 7-14, 2023}, pages 1477--1485, 2023.

\bibitem{cvpr06}
Yiming Li, Zhiding Yu, Christopher~B. Choy, Chaowei Xiao, Jos{\'{e}}~M. {\'{A}}lvarez, Sanja Fidler, Chen Feng, and Anima Anandkumar.
\newblock Voxformer: Sparse voxel transformer for camera-based 3d semantic scene completion.
\newblock In {\em {IEEE/CVF} Conference on Computer Vision and Pattern Recognition, {CVPR} 2023, Vancouver, BC, Canada, June 17-24, 2023}, pages 9087--9098. {IEEE}, 2023.

\bibitem{eccv02}
Zhiqi Li, Wenhai Wang, Hongyang Li, Enze Xie, Chonghao Sima, Tong Lu, Yu Qiao, and Jifeng Dai.
\newblock Bevformer: Learning bird's-eye-view representation from multi-camera images via spatiotemporal transformers.
\newblock In {\em Computer Vision - {ECCV} 2022 - 17th European Conference, Tel Aviv, Israel, October 23-27, 2022, Proceedings, Part {IX}}, volume 13669, pages 1--18. Springer, 2022.

\bibitem{nips02}
Tingting Liang, Hongwei Xie, Kaicheng Yu, Zhongyu Xia, Zhiwei Lin, Yongtao Wang, Tao Tang, Bing Wang, and Zhi Tang.
\newblock Bevfusion: {A} simple and robust lidar-camera fusion framework.
\newblock In {\em Advances in Neural Information Processing Systems 35: Annual Conference on Neural Information Processing Systems 2022, NeurIPS 2022, New Orleans, LA, USA, November 28 - December 9, 2022}, volume~35, pages 10421--10434, 2022.

\bibitem{cvpr12}
Tsung{-}Yi Lin, Piotr Doll{\'{a}}r, Ross~B. Girshick, Kaiming He, Bharath Hariharan, and Serge~J. Belongie.
\newblock Feature pyramid networks for object detection.
\newblock In {\em 2017 {IEEE} Conference on Computer Vision and Pattern Recognition, {CVPR} 2017, Honolulu, HI, USA, July 21-26, 2017}, pages 936--944, 2017.

\bibitem{eccv01}
Yingfei Liu, Tiancai Wang, Xiangyu Zhang, and Jian Sun.
\newblock {PETR:} position embedding transformation for multi-view 3d object detection.
\newblock In {\em Computer Vision - {ECCV} 2022 - 17th European Conference, Tel Aviv, Israel, October 23-27, 2022, Proceedings, Part {XXVII}}, volume 13687, pages 531--548, 2022.

\bibitem{iccv05}
Yingfei Liu, Junjie Yan, Fan Jia, Shuailin Li, Aqi Gao, Tiancai Wang, and Xiangyu Zhang.
\newblock Petrv2: {A} unified framework for 3d perception from multi-camera images.
\newblock In {\em {IEEE/CVF} International Conference on Computer Vision, {ICCV} 2023, Paris, France, October 1-6, 2023}, pages 3239--3249, 2023.

\bibitem{icra01}
Zhijian Liu, Haotian Tang, Alexander Amini, Xinyu Yang, Huizi Mao, Daniela~L. Rus, and Song Han.
\newblock Bevfusion: Multi-task multi-sensor fusion with unified bird's-eye view representation.
\newblock In {\em {IEEE} International Conference on Robotics and Automation, {ICRA} 2023, London, UK, May 29 - June 2, 2023}, pages 2774--2781, 2023.

\bibitem{eccv05}
Jiachen Lu, Zheyuan Zhou, Xiatian Zhu, Hang Xu, and Li Zhang.
\newblock Learning ego 3d representation as ray tracing.
\newblock In {\em Computer Vision - {ECCV} 2022 - 17th European Conference, Tel Aviv, Israel, October 23-27, 2022, Proceedings, Part {XXVI}}, volume 13686, pages 129--144, 2022.

\bibitem{arxiv04}
Ruihang Miao, Weizhou Liu, Mingrui Chen, Zheng Gong, Weixin Xu, Chen Hu, and Shuchang Zhou.
\newblock Occdepth: {A} depth-aware method for 3d semantic scene completion.
\newblock {\em CoRR}, abs/2302.13540, 2023.

\bibitem{arxiv08}
Zhenxing Ming, Julie~Stephany Berrio, Mao Shan, and Stewart Worrall.
\newblock Inversematrixvt3d: An efficient projection matrix-based approach for 3d occupancy prediction.
\newblock {\em CoRR}, abs/2401.12422, 2024.

\bibitem{arxiv02}
Zhenxing Ming, Julie~Stephany Berrio, Mao Shan, and Stewart Worrall.
\newblock Occfusion: {A} straightforward and effective multi-sensor fusion framework for 3d occupancy prediction.
\newblock {\em CoRR}, abs/2403.01644, 2024.

\bibitem{iccv04}
Yair Movshovitz{-}Attias, Alexander Toshev, Thomas~K. Leung, Sergey Ioffe, and Saurabh Singh.
\newblock No fuss distance metric learning using proxies.
\newblock In {\em {IEEE} International Conference on Computer Vision, {ICCV} 2017, Venice, Italy, October 22-29, 2017}, pages 360--368, 2017.

\bibitem{iral01}
Jingyi Pan, Zipeng Wang, and Lin Wang.
\newblock Co-occ: Coupling explicit feature fusion with volume rendering regularization for multi-modal 3d semantic occupancy prediction.
\newblock {\em {IEEE} Robotics Autom. Lett.}, 9(6):5687--5694, 2024.

\bibitem{icra02}
Mingjie Pan, Jiaming Liu, Renrui Zhang, Peixiang Huang, Xiaoqi Li, Hongwei Xie, Bing Wang, Li Liu, and Shanghang Zhang.
\newblock Renderocc: Vision-centric 3d occupancy prediction with 2d rendering supervision.
\newblock In {\em {IEEE} International Conference on Robotics and Automation, {ICRA} 2024, Yokohama, Japan, May 13-17, 2024}, pages 12404--12411, 2024.

\bibitem{cvpr03}
Xuran Pan, Zhuofan Xia, Shiji Song, Li~Erran Li, and Gao Huang.
\newblock 3d object detection with pointformer.
\newblock In {\em {IEEE} Conference on Computer Vision and Pattern Recognition, {CVPR} 2021, virtual, June 19-25, 2021}, pages 7463--7472, 2021.

\bibitem{iclr01}
Jinhyung Park, Chenfeng Xu, Shijia Yang, Kurt Keutzer, Kris~M. Kitani, Masayoshi Tomizuka, and Wei Zhan.
\newblock Time will tell: New outlooks and {A} baseline for temporal multi-view 3d object detection.
\newblock In {\em The Eleventh International Conference on Learning Representations, {ICLR} 2023, Kigali, Rwanda, May 1-5, 2023}, 2023.

\bibitem{eccv03}
Jonah Philion and Sanja Fidler.
\newblock Lift, splat, shoot: Encoding images from arbitrary camera rigs by implicitly unprojecting to 3d.
\newblock In {\em Computer Vision - {ECCV} 2020 - 16th European Conference, Glasgow, UK, August 23-28, 2020, Proceedings, Part {XIV}}, volume 12359, pages 194--210. Springer, 2020.

\bibitem{nips04}
Charles~Ruizhongtai Qi, Li Yi, Hao Su, and Leonidas~J. Guibas.
\newblock Pointnet++: Deep hierarchical feature learning on point sets in a metric space.
\newblock In {\em Advances in Neural Information Processing Systems 30: Annual Conference on Neural Information Processing Systems 2017, December 4-9, 2017, Long Beach, CA, {USA}}, pages 5099--5108, 2017.

\bibitem{cvpr07}
Cody Reading, Ali Harakeh, Julia Chae, and Steven~L. Waslander.
\newblock Categorical depth distribution network for monocular 3d object detection.
\newblock In {\em {IEEE} Conference on Computer Vision and Pattern Recognition, {CVPR} 2021, virtual, June 19-25, 2021}, pages 8555--8564, 2021.

\bibitem{3DV01}
Luis Rold{\~{a}}o, Raoul de Charette, and Anne Verroust{-}Blondet.
\newblock Lmscnet: Lightweight multiscale 3d semantic completion.
\newblock In {\em 8th International Conference on 3D Vision, 3DV 2020, Virtual Event, Japan, November 25-28, 2020}, pages 111--119, 2020.

\bibitem{cvpr08}
Shuran Song, Fisher Yu, Andy Zeng, Angel~X. Chang, Manolis Savva, and Thomas~A. Funkhouser.
\newblock Semantic scene completion from a single depth image.
\newblock In {\em 2017 {IEEE} Conference on Computer Vision and Pattern Recognition, {CVPR} 2017, Honolulu, HI, USA, July 21-26, 2017}, pages 190--198, 2017.

\bibitem{nips01}
Xiaoyu Tian, Tao Jiang, Longfei Yun, Yucheng Mao, Huitong Yang, Yue Wang, Yilun Wang, and Hang Zhao.
\newblock Occ3d: {A} large-scale 3d occupancy prediction benchmark for autonomous driving.
\newblock In {\em Advances in Neural Information Processing Systems 36: Annual Conference on Neural Information Processing Systems 2023, NeurIPS 2023, New Orleans, LA, USA, December 10 - 16, 2023}, volume~36, pages 64318--64330, 2023.

\bibitem{nips03}
Ashish Vaswani, Noam Shazeer, Niki Parmar, Jakob Uszkoreit, Llion Jones, Aidan~N. Gomez, Lukasz Kaiser, and Illia Polosukhin.
\newblock Attention is all you need.
\newblock In {\em Advances in Neural Information Processing Systems 30: Annual Conference on Neural Information Processing Systems 2017, December 4-9, 2017, Long Beach, CA, {USA}}, pages 5998--6008, 2017.

\bibitem{arxiv03}
Guoqing Wang, Zhongdao Wang, Pin Tang, Jilai Zheng, Xiangxuan Ren, Bailan Feng, and Chao Ma.
\newblock Occgen: Generative multi-modal 3d occupancy prediction for autonomous driving.
\newblock {\em CoRR}, abs/2404.15014, 2024.

\bibitem{iccv01}
Xiaofeng Wang, Zheng Zhu, Wenbo Xu, Yunpeng Zhang, Yi Wei, Xu Chi, Yun Ye, Dalong Du, Jiwen Lu, and Xingang Wang.
\newblock Openoccupancy: {A} large scale benchmark for surrounding semantic occupancy perception.
\newblock In {\em {IEEE/CVF} International Conference on Computer Vision, {ICCV} 2023, Paris, France, October 1-6, 2023}, pages 17804--17813, 2023.

\bibitem{cvpr04}
Yuqi Wang, Yuntao Chen, Xingyu Liao, Lue Fan, and Zhaoxiang Zhang.
\newblock Panoocc: Unified occupancy representation for camera-based 3d panoptic segmentation.
\newblock {\em CoRR}, abs/2306.10013, 2023.

\bibitem{cvpr10}
Yuqi Wang, Yuntao Chen, and Zhaoxiang Zhang.
\newblock Frustumformer: Adaptive instance-aware resampling for multi-view 3d detection.
\newblock In {\em {IEEE/CVF} Conference on Computer Vision and Pattern Recognition, {CVPR} 2023, Vancouver, BC, Canada, June 17-24, 2023}, pages 5096--5105, 2023.

\bibitem{corl01}
Yue Wang, Vitor Guizilini, Tianyuan Zhang, Yilun Wang, Hang Zhao, and Justin Solomon.
\newblock {DETR3D:} 3d object detection from multi-view images via 3d-to-2d queries.
\newblock In {\em Conference on Robot Learning, 8-11 November 2021, London, {UK}}, volume 164, pages 180--191, 2021.

\bibitem{pami01}
Xiu{-}Shen Wei, Yi{-}Zhe Song, Oisin~Mac Aodha, Jianxin Wu, Yuxin Peng, Jinhui Tang, Jian Yang, and Serge~J. Belongie.
\newblock Fine-grained image analysis with deep learning: {A} survey.
\newblock {\em {IEEE} Trans. Pattern Anal. Mach. Intell.}, 44(12):8927--8948, 2022.

\bibitem{iccv02}
Yi Wei, Linqing Zhao, Wenzhao Zheng, Zheng Zhu, Jie Zhou, and Jiwen Lu.
\newblock Surroundocc: Multi-camera 3d occupancy prediction for autonomous driving.
\newblock In {\em {IEEE/CVF} International Conference on Computer Vision, {ICCV} 2023, Paris, France, October 1-6, 2023}, pages 21672--21683, 2023.

\bibitem{arxiv01}
Enze Xie, Zhiding Yu, Daquan Zhou, Jonah Philion, Anima Anandkumar, Sanja Fidler, Ping Luo, and Jos{\'{e}}~M. {\'{A}}lvarez.
\newblock M\({}^{\mbox{2}}\)bev: Multi-camera joint 3d detection and segmentation with unified birds-eye view representation.
\newblock {\em CoRR}, abs/2204.05088, 2022.

\bibitem{aaai04}
Xu Yan, Jiantao Gao, Jie Li, Ruimao Zhang, Zhen Li, Rui Huang, and Shuguang Cui.
\newblock Sparse single sweep lidar point cloud segmentation via learning contextual shape priors from scene completion.
\newblock In {\em Thirty-Fifth {AAAI} Conference on Artificial Intelligence, {AAAI} 2021, Thirty-Third Conference on Innovative Applications of Artificial Intelligence, {IAAI} 2021, The Eleventh Symposium on Educational Advances in Artificial Intelligence, {EAAI} 2021, Virtual Event, February 2-9, 2021}, pages 3101--3109, 2021.

\bibitem{sensors01}
Yan Yan, Yuxing Mao, and Bo Li.
\newblock {SECOND:} sparsely embedded convolutional detection.
\newblock {\em Sensors}, 18(10):3337, 2018.

\bibitem{iccv03}
Yunpeng Zhang, Zheng Zhu, and Dalong Du.
\newblock Occformer: Dual-path transformer for vision-based 3d semantic occupancy prediction.
\newblock In {\em {IEEE/CVF} International Conference on Computer Vision, {ICCV} 2023, Paris, France, October 1-6, 2023}, pages 9399--9409, 2023.

\bibitem{cvpr14}
Yin Zhou and Oncel Tuzel.
\newblock Voxelnet: End-to-end learning for point cloud based 3d object detection.
\newblock In {\em 2018 {IEEE} Conference on Computer Vision and Pattern Recognition, {CVPR} 2018, Salt Lake City, UT, USA, June 18-22, 2018}, pages 4490--4499, 2018.

\bibitem{iclr02}
Xizhou Zhu, Weijie Su, Lewei Lu, Bin Li, Xiaogang Wang, and Jifeng Dai.
\newblock Deformable {DETR:} deformable transformers for end-to-end object detection.
\newblock In {\em 9th International Conference on Learning Representations, {ICLR} 2021, Virtual Event, Austria, May 3-7, 2021}, 2021.

\bibitem{arxiv07}
Sicheng Zuo, Wenzhao Zheng, Yuanhui Huang, Jie Zhou, and Jiwen Lu.
\newblock Pointocc: Cylindrical tri-perspective view for point-based 3d semantic occupancy prediction.
\newblock {\em CoRR}, abs/2308.16896, 2023.

\end{thebibliography}
}

\end{document}